\documentclass{article}
\usepackage{amssymb}
\usepackage{soul}

\usepackage[utf8]{inputenc} 
\usepackage[T1]{fontenc}    
\usepackage{url}            
\usepackage{booktabs}       
\usepackage{nicefrac}       
\usepackage{microtype}      
\usepackage{graphics,graphicx,color}

\usepackage{algorithm}
\usepackage{algorithmic}
\usepackage{enumitem}

\usepackage{amsfonts}       
\usepackage{amsmath}       
\usepackage{amssymb}

\newcommand{\Figref}[1]{Figure~\ref{#1}}  
\newcommand{\figref}[1]{Fig.~\ref{#1}}    

\newcommand{\tabref}[1]{Table~\ref{#1}}

\newcommand{\eqnref}[1]{Eq.~\ref{#1}} 
\newcommand{\secref}[1]{Sec.~\ref{#1}} 
\newcommand{\suppref}[1]{Appx.~\ref{#1}}

\usepackage{xspace}
\makeatletter
\DeclareRobustCommand\onedot{\futurelet\@let@token\@onedot}
\def\@onedot{\ifx\@let@token.\else.\null\fi\xspace}
\makeatother

\usepackage{xr-hyper}
\makeatletter
\newcommand*{\addFileDependency}[1]{
  \typeout{(#1)}
  \@addtofilelist{#1}
  \IfFileExists{#1}{}{\typeout{No file #1.}}
}
\makeatother

\usepackage{xcolor}
\usepackage[normalem]{ulem}

\definecolor{ourorange}{HTML}{e19c24}
\definecolor{ourgreen}{HTML}{97b567}
\definecolor{ourred}{HTML}{ec6235}
\definecolor{ourblue}{HTML}{5e81b5}
\definecolor{ourgrey}{HTML}{919191}

\usepackage{subcaption}

\DeclareMathAlphabet{\mathsfit}{\encodingdefault}{\sfdefault}{m}{sl}
\SetMathAlphabet{\mathsfit}{bold}{\encodingdefault}{\sfdefault}{bx}{n}

\newcommand{\norm}[1]{\left\lVert#1\right\rVert}

\newcommand{\blue}[1]{\color{black}{#1}\color{black}} %

\usepackage{multicol}
\usepackage[bookmarks=true]{hyperref}
\usepackage{graphicx}
\usepackage{amsmath}
\usepackage{relsize}
\usepackage{caption}
\usepackage{subcaption}
\usepackage[final]{corl_2025} 

\title{Motion Priors Reimagined: Adapting Flat-Terrain Skills for Complex Quadruped Mobility}

\author{
  Zewei Zhang\\
  Department of Mechanical Engineering, EPFL\\
  \And
  Chenhao Li \\
  ETH AI Center, 
  ETH Zurich \\
  \And
  Takahiro Miki \\
  Robotic Systems Lab,
  ETH Zurich \\
  \And
  Marco Hutter \\
  Robotic Systems Lab,
  ETH Zurich \\
}

\begin{document}
\maketitle

\vspace{-0.7cm}
\begin{abstract}
Reinforcement learning (RL)-based motion imitation methods trained on demonstration data can effectively learn natural and expressive motions with minimal reward engineering but often struggle to generalize to novel environments.
We address this by proposing a hierarchical RL framework in which a low-level policy is first pre-trained to imitate animal motions on flat ground, thereby establishing motion priors.
A subsequent high-level, goal-conditioned policy then builds on these priors, learning residual corrections that enable perceptive locomotion, local obstacle avoidance, and goal-directed navigation across diverse and rugged terrains.
Simulation experiments illustrate the effectiveness of learned residuals in adapting to progressively challenging uneven terrains while still preserving the locomotion characteristics provided by the motion priors.
Furthermore, our results demonstrate improvements in motion regularization over baseline models trained without motion priors under similar reward setups. 
Real-world experiments with an ANYmal-D quadruped robot confirm our policy’s capability to generalize animal-like locomotion skills to complex terrains, demonstrating smooth and efficient locomotion and local navigation performance amidst challenging terrains with obstacles.
\end{abstract}

\vspace{-0.3cm}

\begin{figure}[h!]
    \centering
    \includegraphics[width=0.90\textwidth]{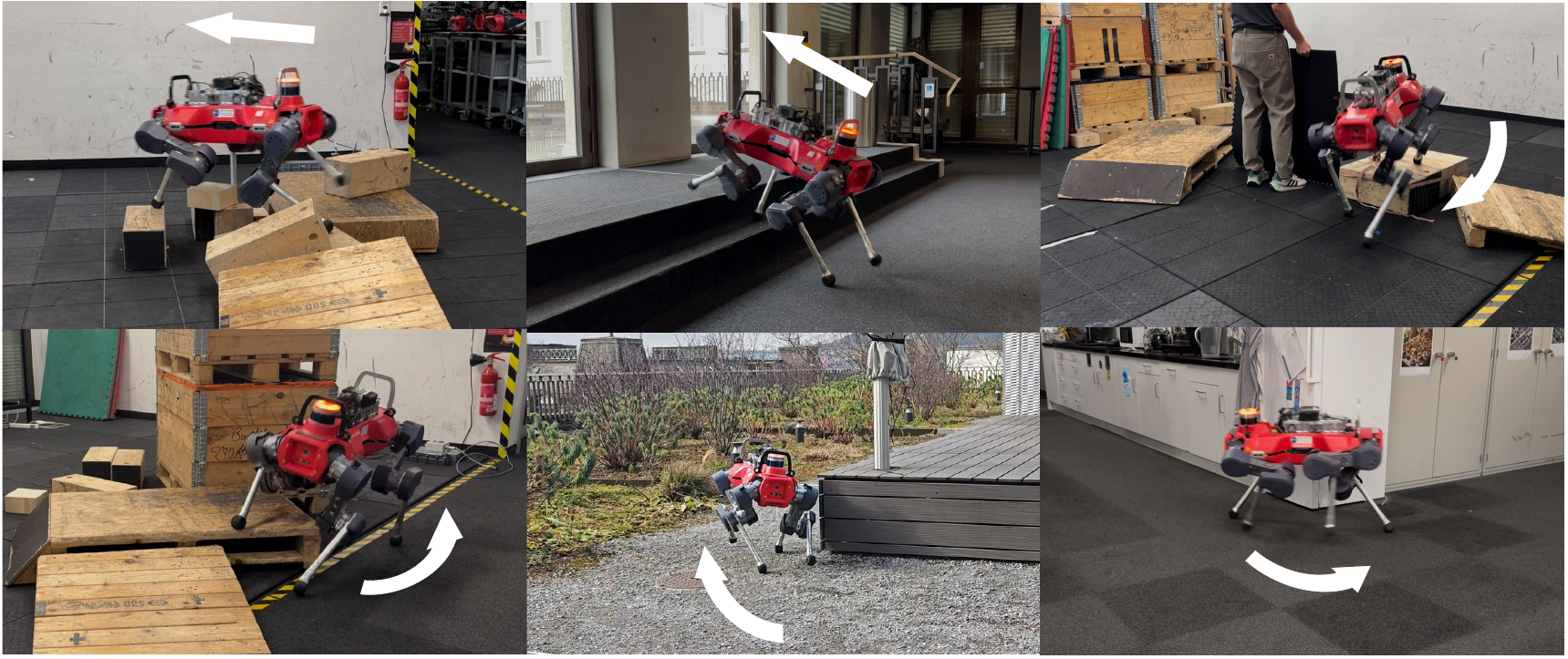}
    \caption{
    ANYmal-D hardware experiments in indoor/outdoor terrains (stairs, random blocks, high obstacles). The white arrow marks the direction of movement toward the specified goal position. The robot employs an animal-like gait to traverse uneven ground and avoid obstacles, demonstrating our policy’s transfer of flat-ground motion priors to complex environments. (See supplementary videos at \url{https://anymalprior.github.io/})
    }
    \label{fig:result_cover}
    \vspace{-0.6cm}
\end{figure}
\section{Introduction} \label{sec:intro}
\vspace{-0.2cm}
Legged locomotion remains one of the most challenging problems in robotics, and reinforcement learning (RL) has recently emerged as a promising approach to tackle this complexity.
Contemporary RL locomotion controllers have successfully enabled robots to achieve smooth and stable motion while accurately tracking base velocity~\citep{doi:10.1126/scirobotics.aau5872, doi:10.1126/scirobotics.abk2822,zhuang2024humanoid} or goal position commands~\cite{cheng2023parkour,zhang2024learningagilelocomotionrisky,ren2025vbcomlearningvisionblindcomposite} across various terrains.
Additionally, position-based locomotion policies have demonstrated the capability to perform effective local navigation without relying on an external high-level planning module.
Despite these advancements and their successful deployment on hardware platforms, both velocity-based and position-based locomotion controllers heavily depend on meticulous reward design and struggle to learn expressive, animal- or human-like motions.  

Motion imitation has emerged as an efficient alternative to alleviate these challenges by guiding RL with reference trajectories obtained from teleoperation~\citep{fu2024mobile, zhao2023learningfinegrainedbimanualmanipulation}, motion capture~\citep{he2025asap, Han_2024}, or trajectory optimization~\citep{vollenweider2022advancedskillsmultipleadversarial}. 
Such motion imitation-based policies have enabled natural and agile maneuvers, such as jumping~\citep{Han_2024, li2023versatile} and backflips~\citep{li2023learning}, which is particularly challenging to learn from scratch due to the extensive reward design and tuning.

However, motion imitation-based method is not without its limitations. 
Their performance is inherently bounded by the fidelity and specificity of the reference data.
When a policy is strictly forced to mimic a fixed set of demonstrations, it tends to overfit to the specific characteristics how the reference data are recorded. 
As a result, when deployed in an environment with different dynamics (e.g. terrain types), the policy often fails to reproduce the intended motion style consistently.
This issue arises due to covariate shift, the mismatch between the distribution of states encountered during training (from the demonstration data) and those encountered in the new environment, which can lead to degraded performance and style inconsistency.
Although recent works have sought to improve adaptation beyond the environment where demonstration data are collected~\citep{Han_2024,wu2023RAL,smith2023learningadaptingagilelocomotion}, these approaches either rely on reference trajectories through trajectory optimization method or are confined to setups with only minimal environmental differences. 
Adaptation in more challenging and diverse scenarios using readily accessible, minimally processed references, such as raw retargeted animal motion data, remains underexplored.

In this work, we propose a hierarchical reinforcement learning pipeline that addresses motion imitation challenges.
Our framework begins by pre-training motion priors using motion imitation-based RL on animal datasets collected exclusively from flat terrain. 
These priors subsequently facilitate the training of a high-level goal-reaching policy within a position-based formulation, enabling effective adaptation to complex, non-flat environments and accomplishing perceptive locomotion and local navigation.
During deployment, our policy navigates to target goals while preserving the natural animal-like motion style and executing smooth obstacle avoidance behaviors.

Our contributions are: \textbf{(a)} a hierarchical RL framework that combines pre-training of motion priors from flat-terrain animal data with a task training stage that learns residual corrections for improved adaptability; \textbf{(b)} a unified pipeline achieving both perceptive locomotion and local navigation in a smooth, animal-like gait; and \textbf{(c)} a comprehensive analysis of how motion priors and learned residuals influence overall task and locomotion performance.

Simulation and hardware experiments demonstrate that our framework not only retains the training efficiency of motion imitation but also effectively generalizes to challenging, real-world locomotion and local navigation tasks.
\vspace{-0.3cm}


\section{Related Work} \label{sec:related_work}
\vspace{-0.2cm}
\subsection{Motion Imitation with Reinforcement Learning}
\vspace{-0.2cm}
Motion imitation via RL has emerged as a reliable strategy for acquiring diverse motor skills directly from reference data. 
In legged locomotion, such approaches have been effective at transferring demonstrative behaviors to real-world systems.
For instance, \citet{RoboImitationPeng20} extend Deepmimic framework~\citep{2018-TOG-deepMimic} to real quadrupeds by using retargeted animal motion data and reward defined by the tracking error between the reference and actual states.
Similarly, \citet{he2025asap} have shown that humanoid robots can achieve agile, long-range movements leveraging comparable techniques.
Complementary studies by \citet{li2024fld} and \citet{watanabe2025dfmdeepfouriermimic} further streamline the imitation pipeline by automating motion representation and phase labeling for Deepmimic-based methods.  

An alternative line of research employs adversarial motion priors (AMP), where a GAN-style discriminator learns the distribution of state trajectories from demonstration data~\citep{li2023versatile,li2023learning,2021-TOG-AMP,Escontrela22arXiv_AMP_in_real}. 
Nevertheless, AMP-based approaches still face inherent adversarial learning challenges, including instability during discriminator training and mode collapse, especially when the dataset contains diverse motions~\citep{Peng:EECS-2021-267}. 
In our work, we adopt the Deepmimic-based approach from \citet{li2024fld} due to its ability to yield refined motion representations and training tractability compared to the AMP-based methods.
\vspace{-0.2cm}
\subsection{Task Adaptation with Motion Priors}
\vspace{-0.2cm}
Motion imitation-based RL policies often serve as low-level motion priors that underpin more complex, high-level tasks.
These motion skills can be encoded into a low-dimensional latent space and reused by high-level task policies through hierarchical conditioning~\citep{Han_2024,luo2024universal,2022-TOG-ASE}.
Alternatively, some methods directly incorporate style rewards into task training~\citep{wu2023RAL,Escontrela22arXiv_AMP_in_real}.

For example, \citet{Han_2024} propose a two-stage training strategy where a low-level imitation policy based on animal motions is first acquired and then used to train high-level policies that adapt to various tasks on rough terrains. 
Although their method successfully generates animal-like motions across multiple tasks on various terrains, it remains dependent on training with motion data from uneven terrains and on manually calibrating the simulation environment to match the specific environmental setups under which the training data are collected for the imitation performance.

In contrast, \citet{wu2023RAL} achieve stable locomotion on complex terrains using motion priors learned solely from flat-terrain data, by integrating an AMP-based reward into the task training.
However, this approach relies on carefully crafted trajectories obtained via trajectory optimization algorithms and typically captures a trotting gait.
Extending the generalization of motion priors derived from raw retargetted animal data, including other gait patterns such as walking, pacing, or cantering into novel terrains, remains underexplored.
Our approach addresses this gap by learning joint residuals on top of low-level priors which is only trained on flat terrain, thereby enhancing adaptability to diverse and challenging environments beyond where the motion data are recorded.

\begin{figure*}[t]
    \centering
    \includegraphics[width=0.95\textwidth]{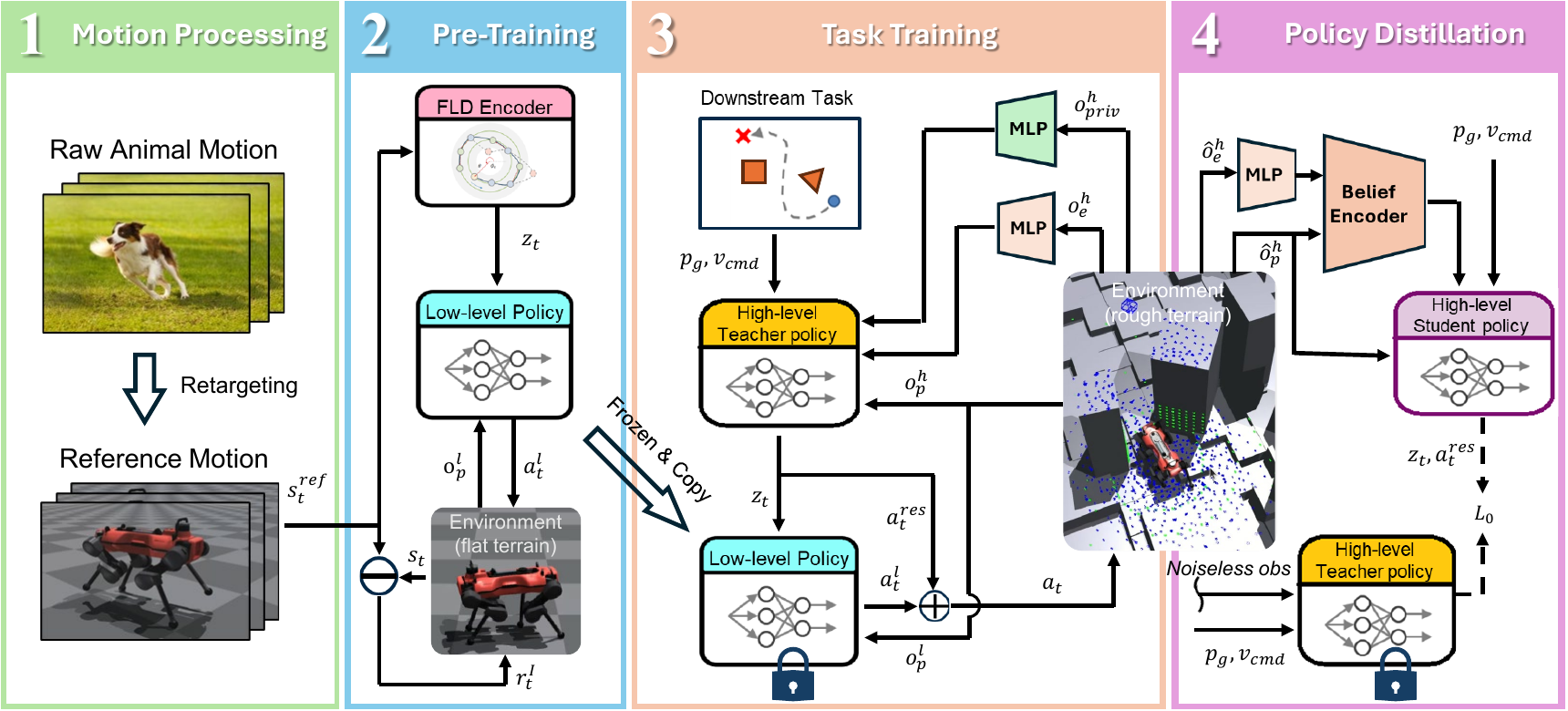}
    \caption{
    Overview of our training framework for quadruped locomotion and local navigation: (1) retarget animal mocap data based on the robot’s configuration;(2) pre-train an FLD encoder and low-level policy on flat terrain for motion priors; (3) train a high-level teacher to output latent commands and joint residuals for rough-terrain adaptation; (4) distill into a student policy for sim-to-real using only noisy observations. Green dots: downsampled Velodyne-LiDAR scans; blue dots: elevation scans around the robot's feet.
    }
    \label{fig:framework}
    \vspace{-0.55cm}
\end{figure*}
\vspace{-0.2cm}
\subsection{Local Navigation at Locomotion Level}
\vspace{-0.2cm}
In autonomous legged robotics, local navigation is commonly managed by a high-level planning module that continuously outputs waypoints~\citep{anymalparkour,lorenz2021navigation,yang2023iplannerimperativepathplanning} or velocity commands~\citep{zhang2024navigation,lee2024navigation} for a low-level locomotion controller.
However, developing an integrated, end-to-end policy that simultaneously handles locomotion and navigation is desirable, as it leverages the full capabilities of the robot to determine optimal actions.

Prior work using position-based task rewards~\citep{ren2025vbcomlearningvisionblindcomposite,rudin2022advancedskillslearninglocomotion} has demonstrated emergent local navigation behavior within a single learned locomotion policy.
Yet, these approaches generally necessitate extensive reward shaping to produce smooth motions and are predominantly applied to local navigation tasks in near-flat terrains.

Building on these developments, our aim is to improve the efficiency and performance of local navigation training at the locomotion level.
By integrating motion priors derived from animal demonstrations (originally collected on flat ground) into a position-based locomotion framework, our pipeline produces a smooth, animal-like gait that empowers the robot to traverse varied terrains and efficiently circumvent local obstacles en route to its goal.
\vspace{-0.2cm}
	

\section{Method} \label{sec:method}
\vspace{-0.2cm}
In this section, we describe our hierarchical RL training framework, which is designed to harness the benefits of motion imitation and train a more powerful high-level policy to extend the capabilities of the original motions only feasible on flat ground for the acquisition of locomotion, local obstacle avoidance, and goal-reaching navigation skills on more challenging terrains. 

As illustrated in \figref{fig:framework}, our approach is divided into four phases: motion processing, pre-training, task training and policy distillation for sim2real. 
Prior to pre-training, animal motion data are retargetted to align with the robot's configuration. 
During pre-training, we first train the FLD encoder on offline data and subsequently learn a low-level policy using the extracted latent representations as the input to generate physical motions.
In the task training phase, the frozen low-level policy acts as a motion prior, while we train a high-level teacher policy with privileged information to handle complex tasks over diverse terrains by outputting latent command to the low-level policy and augmenting the low-level skills with residuals. 
Lastly, the high-level teacher policy is distilled into the student policy with noisy observation to narrow the sim2real gap.
The following subsections provide further details.
\vspace{-0.2cm}
\subsection{Motion Preprocessing}
\vspace{-0.2cm}
Before training the policy to mimic animal movements, the raw motion capture data need to be retargetted to match the robot’s kinematic configuration. 
We employ inverse kinematics to convert the raw data based on the positions of the neck, pelvis, and feet, following the established pipeline in~\citet{RoboImitationPeng20}. 
The selected animal mocap data, sourced from~\citep{Han_2024,mann}, comprise various gait patterns, such as walking, pacing, and cantering, performed at different speeds on flat ground.
\vspace{-0.2cm}
\subsection{Pre-training} \label{sec:fld_model_and_low_level}
\vspace{-0.2cm}
Given the retargeted motion data, the pre-training stage focuses on acquiring a low-level policy that mimics the demonstrated motor skills.
Crucial to this process is the training of an encoder that compresses the motion trajectories into low-dimensional latent vectors, thereby enabling efficient reuse of these low-level skills in later stages.

We first pre-train the FLD model which comprises both an encoder and a decoder, to capture a representation of motion patterns before proceeding to train the low-level imitation policy, following the framework outlined in~\citet{li2024fld}.
More information on FLD training is detailed in \suppref{sec:detail_fld}.

With the refined motion representation in hand, we next leverage the latent embedding to train a low-level motion imitation policy.
In this phase, the policy is trained on flat terrain while adhering to the FLD pipeline for motion learning. However, several modifications are introduced to boost imitation performance.
Specifically, we eliminate the decoder and replace the reconstructed trajectories with ground-truth trajectories, bypassing the phase propagation mechanism entirely.
During training, motion clips are randomly sampled from the dataset, and the encoder dynamically computes latent embeddings from the input reference state sequences.
These modifications may help reduce performance degradation from FLD decoder reconstruction errors while still maintaining a structured latent space that captures the motion’s global periodic features.

The low-level policy operates in an action space defined by 12-dimensional joint actions $a_t^l$, and its observation $o_p^l$ comprises proprioceptive signals including base linear and angular velocities, projected gravity vectors, joint positions, and the latent encodings. 
The reward function is composed of an imitation component and a regularization component.
Additional details regarding the reward functions are provided in \suppref{sec:reward_low_high_detail}. 
Consequently, the trained low-level policy that bridges the robot's motor skills with low-dimensional latent embeddings not only serves as the command interface for executing low-level skills, but also lays the foundation for subsequent high-level policy training.

\vspace{-0.2cm}
\subsection{Task Training}
\vspace{-0.2cm}
After training the low-level policy as a motion prior on flat-ground data, we extend its capabilities to complex environments via task training.
In this phase, a high-level teacher policy is learned on top of the frozen low-level policy to address locomotion and local navigation tasks across various terrains. 

The high-level teacher policy outputs 16-dimensional latent commands $z_t$ and 12-dimensional joint residuals $a_t^{res}$ to refine the basic motion skills. 
Its observation space includes noiseless proprioceptive inputs $o_{p}^h$ (identical to those used by the low-level policy), noiseless exteroceptive inputs $o_e^h$, privileged states $o_{priv}^h$ (capturing leg contact information, friction coefficients, and external disturbances), and task-specific inputs.
For exteroception, we fuse elevation scans around each robot foot~\citep{doi:10.1126/scirobotics.abc5986} with a downsampled Velodyne LiDAR scan arranged in a sparse conical pattern to provide a comprehensive environmental profile.
As illustrated in \figref{fig:framework}, we further employ small MLPs to encode these terrain and privileged state inputs.

For the downstream task, our objective is to train an integrated policy that leverages motion priors to reach a goal in challenging, rough terrains including uneven surfaces, stairs, slopes, and high obstacles. 
The task input comprises the goal position $p_g=(p_{g,x},p_{g,y})$, defined relative to the robot's base frame, and a velocity command $v_{cmd}$ that coarsely regulates forward motion toward the goal. 
We define the task rewards as follows:
\vspace{-0.cm}
\begin{equation}
    r_{reach} = \frac{1}{T_r}\left(1-\frac{\norm{\mathbf{d}}_2}{2}\right) \quad \text{if} \; t > T-T_r \; \text{and} \; \norm{\mathbf{d}}_2 < 2; \text{else} \; 0,
    \label{eq:tracking_rew}
\end{equation}
\begin{equation}
    r_{vel} = \min(v_{cmd}, \langle \mathbf{v}, \mathbf{d} \rangle) \quad \text{if} \; \norm{\mathbf{d}}_2 > 0.15; \text{else} \; 0,
    \label{eq:vel_rew}
\end{equation}
where $t$, $T$, and $T_r$ denote the current time, the interval between successive position commands, and the threshold time for reward computation, respectively.
$\mathbf{d}$ denotes the displacement vector from the robot base to the target, and $\mathbf{v}$ represents the robot's base linear velocity. 
Since $r_{reach}$ tends to be sparse, the additional velocity reward $r_{vel}$ provides a denser learning signal to effectively guide local navigation in cluttered environments with high obstacles. 

We maintain the same regularization rewards and weights used in the low-level policy (excluding tracking terms) and add new residual penalty terms to constrain the joint residual corrections (\eqnref{eq:joint_res}).
\begin{equation}
    r_{res} = w_{res} \sum_{i=1}^{12}(a^{res}_{t,i})^2, 
    \label{eq:joint_res}
\end{equation}
where $w_{res}$ denotes the weight for the residual penalty.
More details on the reward functions for the high-level teacher policy are provided in~\suppref{sec:reward_low_high_detail}.

\begin{figure}[t!]
    \centering
    \includegraphics[clip, trim=0cm 0cm 0cm 0cm,width=0.86\textwidth]{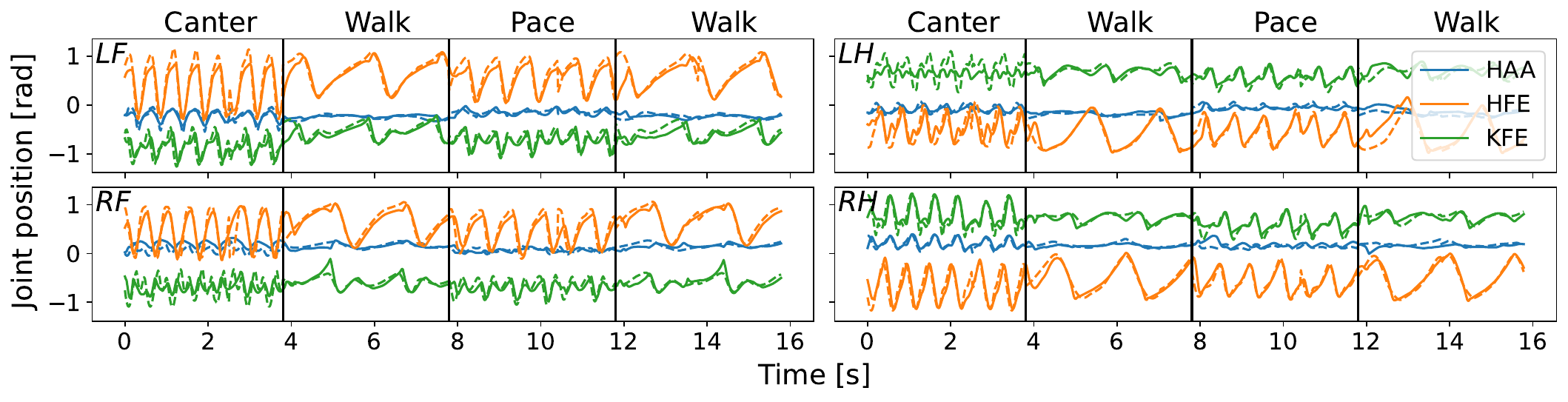}
    \vspace{-0.1cm}
    \caption{
    Actual vs. reference joint positions for the low-level policy on flat ground. Dashed lines are reference trajectories; solid lines are the actual trajectories. The plot (canter → walk → pace → walk) shows close alignment between reference and actual motions.
    }
    \label{fig:dof_pos_transition}
    \vspace{-0.45cm}
\end{figure}
\vspace{-0.2cm}
\subsection{Policy Distillation for Sim2Real}
\vspace{-0.2cm}
To bridge the sim-to-real gap in perceptive locomotion, we employ a privileged learning strategy~\citep{pmlr-v100-chen20a} to distill a high-performance teacher policy into a student policy.
In our approach, the teacher policy is initially trained under ideal conditions using privileged, noiseless proprioceptive and exteroceptive observations.
Subsequently, we distill a student policy from the teacher that operates solely on noisy proprioceptive and exteroceptive data, without access to privileged information through supervised learning.
As shown in \figref{fig:framework}, a recurrent belief encoder, implemented with a Gated Recurrent Unit (GRU), is integrated into student policy to reconstruct the privileged state and recover noiseless exteroceptive signals from noisy proprioceptive and exteroceptive inputs.
The design of the belief encoder follows~\citet{doi:10.1126/scirobotics.abk2822}. 
The distillation process is guided by a behavior loss $L_0$ that quantifies the discrepancy between the teacher’s and student’s actions.
This teacher-student framework allows us to first establish a high-performance teacher policy under controlled conditions, and then transfer that performance to a student policy that is robust under realistic, noisy operating conditions.
\vspace{-0.3cm}


\section{Experiments} \label{sec:results}
\vspace{-0.2cm}
\subsection{Experiment Setup}
\vspace{-0.2cm}
We select ANYmal-D as our robot platform and introduce several experiments to demonstrate and verify the effectiveness of our framework. 
Our experiments aim to \textbf{(a)} test whether a policy enriched with latent motion priors and joint residual corrections can reliably navigate to target goals and avoid local obstacles on complex terrains; \textbf{(b)} evaluate how different penalties on residual actions affect locomotion and goal-reaching performance; and \textbf{(c)} examine the motion regularization performance improvement over a baseline RL controller trained from scratch under similar reward conditions.
\begin{figure*}[t!]
    \centering
        \includegraphics[clip, trim=0cm 0cm 0cm 0cm,width=0.90\textwidth]{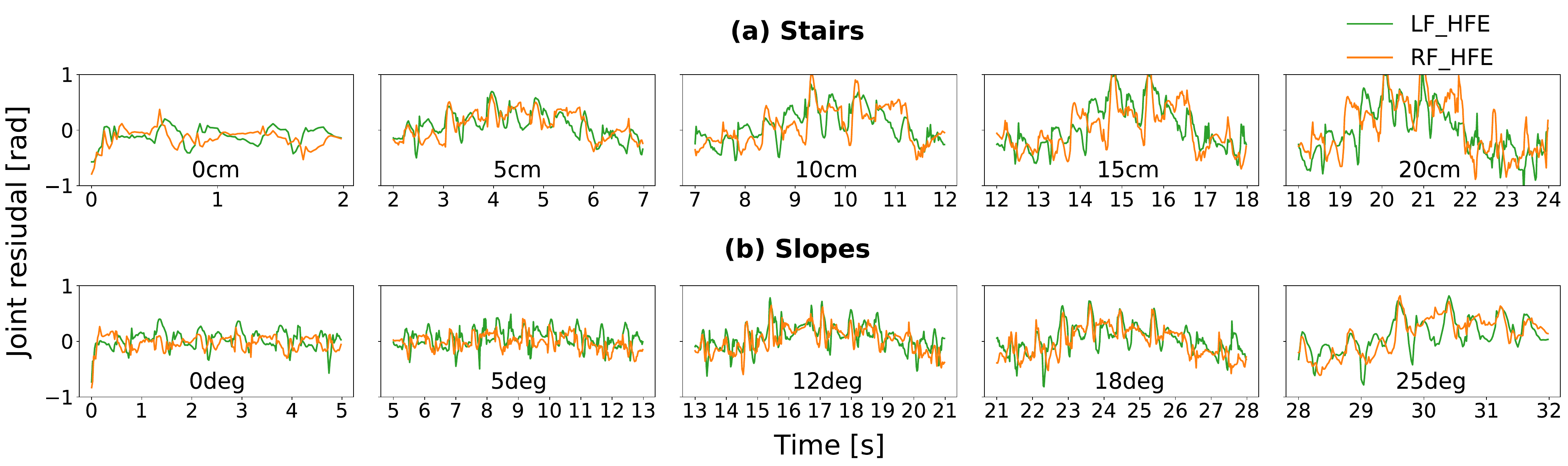 }
        \vspace{-0.1cm}
    \caption{Comparison of high-level joint residuals for the HFE joint on the left front (LF) and right front (RF) leg across two terrain types: (a) pyramid stairs and (b) pyramid slopes with a rugged surface (see \figref{fig:residual_terrain_levels}). Each terrain type is divided into five difficulty levels (displayed at the bottom of each subplot), with difficulty increasing from left (easy) to right (hard).}
    \label{fig:residual_low_level_twoterrains}
    \vspace{-0.2cm}
\end{figure*}
\begin{figure}[t!]
    \centering
    \includegraphics[clip, trim=0cm 0cm 0cm 0cm,width=0.91\textwidth]{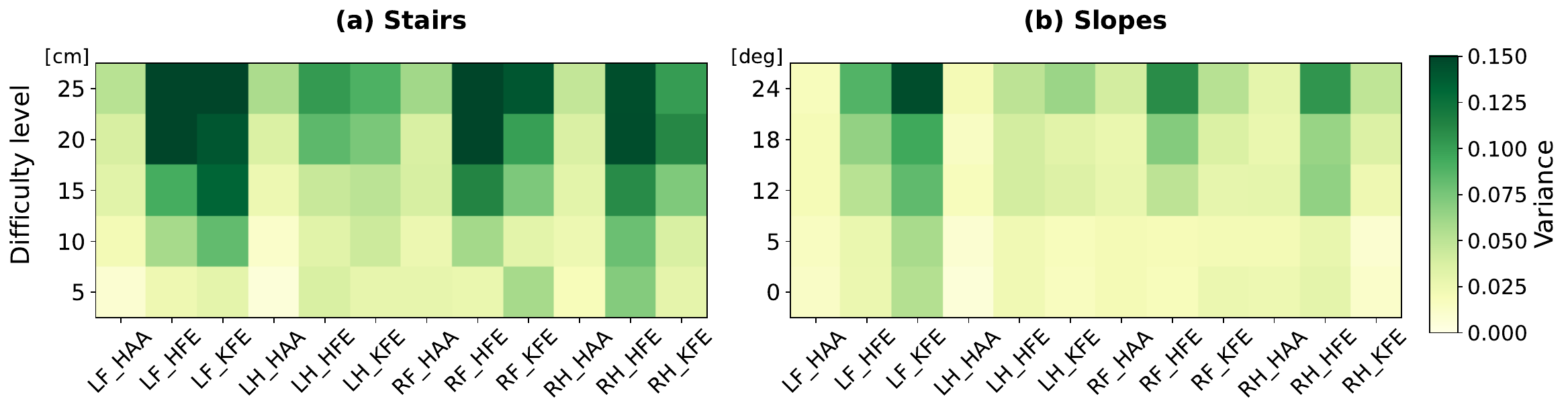}
    \vspace{-0.1cm}
    \caption{Variance in joint residuals of LF-HFE joint, as observed in the experiment depicted in \figref{fig:residual_terrain_levels}. Darker colors denote higher involvement of joint residuals on specific terrain types.}
    \label{fig:var_joint_res}
    \vspace{-0.45cm}
\end{figure}
\vspace{-0.2cm}
\subsection{Simulation Results}
\vspace{-0.2cm}
\subsubsection{Low-level policy}
\vspace{-0.2cm}
We train and test a low-level policy that can demonstrate multiple animal-like motions imitated in simulation. 
As mentioned in \secref{sec:method}, the low-level policy mimics motions that include walking, pacing, and cantering gaits, and each is at different forward velocities. 
Each skill can be performed on flat terrain and also transition smoothly between each other by commanding different latent embeddings. \Figref{fig:dof_pos_transition} highlights the strong imitation performance of the low-level policy by comparing the reference positions with the actual positions of all joints.
\Figref{fig:foot_sequence_low_level} also shows the footfall sequence of individual motion skills learned in the low-level policy. 
\vspace{-0.2cm}
\subsubsection{High-Level Policy} \label{sec:highlevel}
\vspace{-0.2cm}
We test our high-level student policy in simulation under exteroceptive noise conditions that resemble real-world scenarios. 
The evaluation is carried out on multiple terrains, including stairs, boxes of varying heights, slopes with rugged surfaces, and high obstacles. 
Examples of these terrains can be found in \suppref{sec:hyper_low_high_detail}. 
In all terrain types tested, our high-level policy achieves a high success rate in reaching target positions. 
The policy effectively learns to generate appropriate residuals on top of the low-level motions. 
\blue{By integrating the animal-style smoothness derived from low-level motion priors with adaptive high-level residuals, } the robot exhibits natural gait behaviors and bypassed local obstacles directly through perception, eliminating the need for a separate navigation module to output waypoints. 
Refer to \suppref{sec:eval_student_policy_sim} for a detailed analysis of the performance.

As shown in \figref{fig:residual_low_level_twoterrains} and \figref{fig:var_joint_res}, we analyze the residual changes with the increasing difficulty levels of the terrains on the pyramid stairs and the pyramid slopes (\figref{fig:residual_terrain_levels}). 
Based on the results in \figref{fig:var_joint_res}, we observe an increasing trend in the variance of the joint residuals in all joints as the terrain on which the robot is walking becomes more challenging, suggesting a growing contribution of residuals. 
For a detailed analysis, we select the joint residuals of the joint LF-HFE and RF-HFE as a representative. 
According to the plots, the high-level policy continuously adjusts the joint residual based on the current terrain situation. 
Peaks with continuously increasing height can be observed in \figref{fig:residual_low_level_twoterrains} (a) and (b). 
The variation in the joint residual likely reflect changes in the robot's base orientation and the challenges of traversing steps of varying heights or slopes.
Since the low-level motions alone cannot fully address these conditions, an adaptive residual component is needed to maintain effective locomotion.
These results demonstrate that the high-level policy can successfully leverage joint residuals to generalize to non-flat and complex terrains.

\begin{figure}[t!]
    \centering
    \begin{subfigure}[b]{0.4\textwidth}
        \centering
        \includegraphics[width=\textwidth]{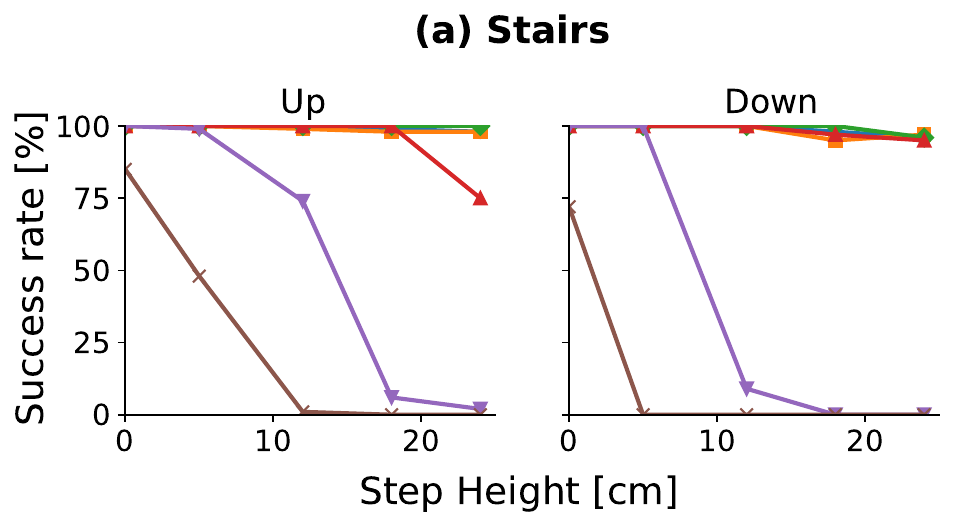}
        \label{fig:stairs_success_rate}
    \end{subfigure}
    \vspace{-0.5cm}
    \begin{subfigure}[b]{0.53\textwidth}
        \centering
        \includegraphics[width=\textwidth]{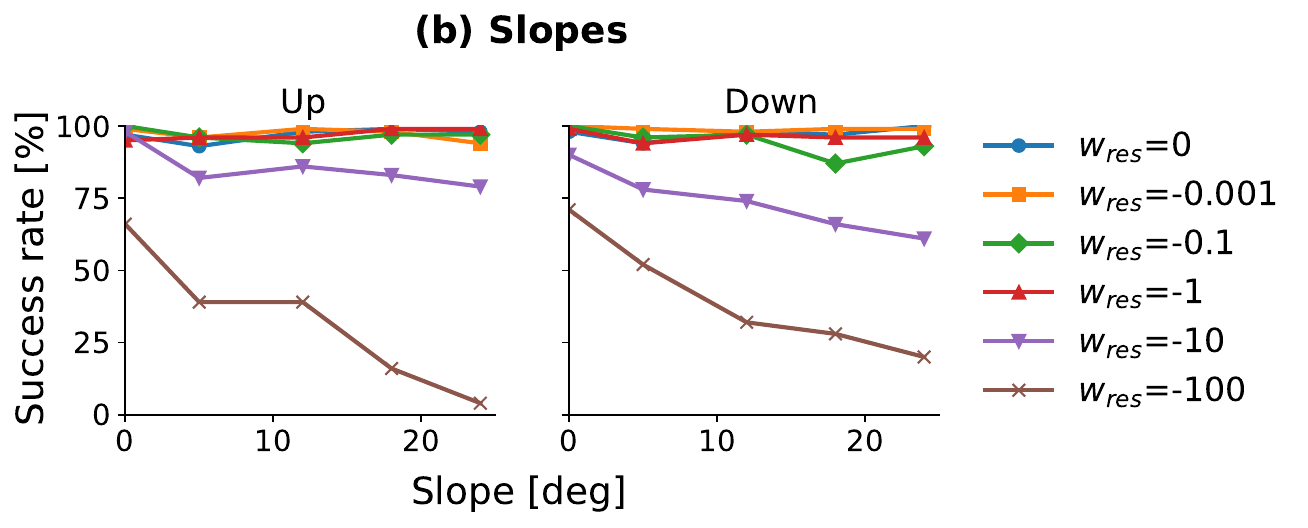}
        \label{fig:slope_success_rate}
    \end{subfigure}
    \vspace{-0.10cm}
    \caption{Goal-reaching success rate on stairs and sloped terrains across different difficulty levels, with the left subplots representing ascent and the right ones representing descent. Each terrain type is evaluated over 100 trials, with randomized initial robot poses in each experiment.}
    \label{fig:success_rate_comparison}
    \vspace{-0.2cm}
\end{figure}

\begin{figure}[t!]
    \centering
    \includegraphics[width=0.65\linewidth]{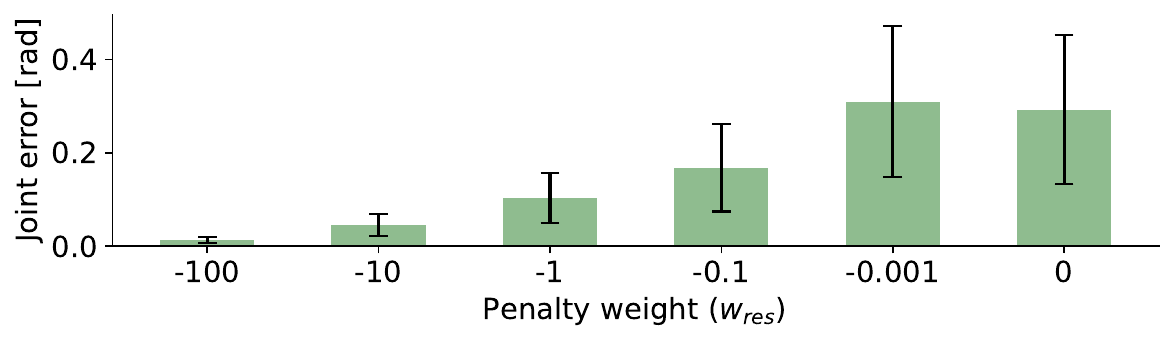}
    \vspace{-0.25cm}
    \caption{Average deviation of actual joint positions from reconstructed reference trajectories with FLD decoder over five seconds of forward walking on flat ground, evaluated at different residual penalty weights. 
    Lower deviations indicate gait behavior that more closely matches the animal motions. }
    \label{fig:error_joint_flat}
    \vspace{-0.45cm}
\end{figure}

\vspace{-0.2cm}
\subsubsection{Impact of Penalty on Residual Actions}
\vspace{-0.2cm}
\blue{We find that training with decreasing values of the joint residual penalty results in stronger adaptation ability but produces less-regularized motions.} 
We train six high-level teacher policies by selecting six penalty weights of the joint residuals ranging from -100 to 0, and investigate how it would affect the performance of locomotion and task completion. 
As shown in \figref{fig:success_rate_comparison}, all policies achieve a high success rate in near-flat terrain. 
However, an excessively high penalty (the weight larger than -10) reduces the efficiency of tackling higher steps and steeper slopes. 
\blue{By contrast an overly low penalty (the weight less than -0.001) may allow residuals to grow without bound, yielding high success rate but resulting in gaits that no longer follow the regularized motions provided by the motion priors, even on flat and near-flat terrain (see~\figref{fig:error_joint_flat} and supplementary videos).
This illustrates a tradeoff between preserving the motion style imparted by the priors and boosting the policy’s adaptability across varied terrains.
}

The performance of our high-level policy underscores the effective regularization achieved by the motion priors embedded in the low-level policy.
Rather than engaging in extensive tuning of multiple reward weights, our approach hinges on a single primary penalty term that constrains the sum of the squared residuals.
Our comparative experiments further confirm that including this penalty is crucial for balancing task-completion rewards with deviations from the motion priors, thereby improving the learned policy’s overall performance.
\vspace{-0.2cm}
\subsubsection{Comparison with Baseline RL Model} \label{sec:com_baseline}
\vspace{-0.2cm}
To evaluate the impact of incorporating motion priors in task-specific training, we compare our high-level teacher policy with a baseline RL model that is trained entirely from scratch without leveraging any low-level motion priors.
Both models are provided with the same observation space configuration except for latent encodings. 
However, the baseline model’s action space is limited to a 12-dimensional vector of joint actions.
\blue{The results display that the baseline policy, trained under the same reward setup without additional tuning and exploration techniques, tends to exhibit a jumping gait despite its ability to navigate challenging terrains (see supplementary videos). 
Incorporating additional regularization terms tuning on base acceleration or vertical velocity, for example, could encourage more natural locomotion. 
In contrast, our model, which integrates motion priors, achieves animal-like gait under identical reward conditions only by including the penalty term for joint residuals. 
This indicates that the learned motion priors inherently enforce the natural movement style learned from animal data, enabling us to cut back on extra regularization and streamline the tuning process. 
}

\begin{figure*}[t!]
    \centering
        \includegraphics[clip, trim=0cm 0cm 0cm 0cm,width=0.98\textwidth]{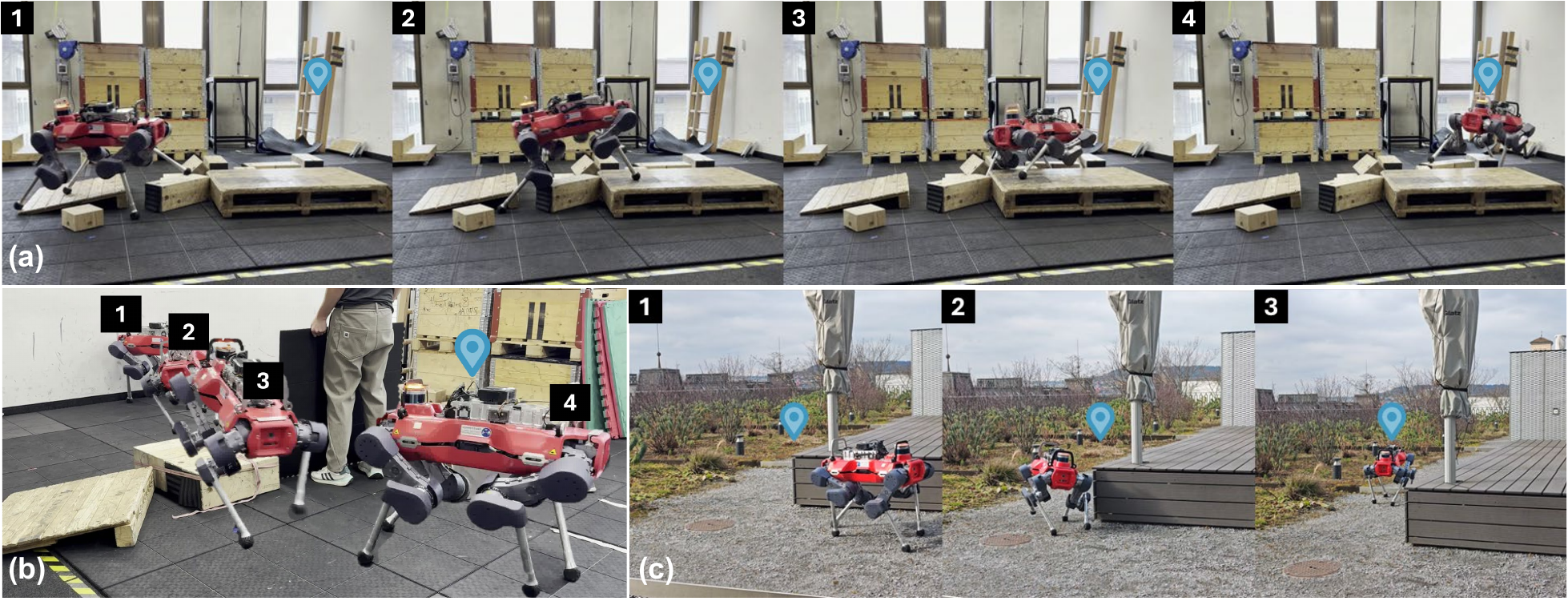 }
    \caption{Real world deployment of the high-level student policy. The sequence of images illustrates the ability of the robot to reach the goal position (blue map pin) on complex terrain while avoiding the local obstacles.
    }
    \label{fig:hardware_exp}
    \vspace{-0.4cm}
\end{figure*}
 
\vspace{-0.2cm}
\subsection{Evaluation in Real World}
\vspace{-0.2cm}
In the final part, we evaluate the high-level student policy on the ANYmal-D robot in real-world environments (see \figref{fig:result_cover} and \figref{fig:hardware_exp}). 
The test area comprises random steps, stairs, and various obstacles that the robot must avoid. 
With a fixed goal provided, the robot navigate to the target while maintaining a smooth animal-style gait without requiring explicit waypoint generation for obstacle bypassing. 
We encourage readers to refer to supplementary videos for further details. 
These results further validate the locomotion and local navigation capabilities of our system in real-world scenarios. 
\vspace{-0.3cm}


\section{Conclusion} 
\label{sec:conclusion}
\vspace{-0.2cm}
We presented a hierarchical RL framework that fused low-level animal motion priors with high-level residual learning for goal-directed locomotion across complex terrains. 
This design reduced reward-tuning effort for motion regularization and improved the robustness and adaptability of flat-terrain skills. 
In simulation, we showed that learned joint residuals achieved a tradeoff between adhering to motion priors and adapting to novel terrains, and that incorporating these priors produced more natural animal-style gaits than baseline RL controllers under similar reward conditions.
Finally, hardware experiments with ANYmal-D confirmed its capability for perceptive locomotion and obstacle-aware local navigation across diverse terrains.



\section{Limitations and Future Work} \label{sec:limitation}
\vspace{-0.2cm}
Despite good performance in perceptive locomotion and local navigation, our approach has several limitations. 
First, our high-level policy can suffer from mode collapse, habitually relying on a single low-level gait and using residuals only to adapt. 
Incorporating exploration strategies or diversity-driving rewards during training may alleviate this issue and encourage the policy to exploit the full range of low-level motion skills.

Second, our training scenarios remain somewhat constrained. 
Future work could be extending our low-level policy to a wider range of motor behaviors (e.g. jumping, crawling) and improving the adaptability for even more challenging terrains, such as gaps, stepping stones, and environments with overhanging obstacles. 

In addition, although this work primarily focuses on active learning to command low-level motions for high-level tasks and demonstrates adaptation with a limited set of motions, we believe the proposed training architecture also offers an efficient and scalable framework for enhancing skill adaptation.
This can be achieved by randomly sampling latent encodings of different motions during training while manually commanding fixed, specific motions for high-level task inference.

\acknowledgments{This research was supported by the ETH AI Center and the Swiss National Science Foundation through the National Centre of Competence in Automation (NCCR Automation). We also thank Hehui Zheng for her assistance with the hardware experiments. }


\bibliography{references}  


\section*{Appendix}
\section{Training Details} 

\subsection{Details for FLD Model Training} \label{sec:detail_fld}
As mentioned in \secref{sec:fld_model_and_low_level}, we train FLD model and use the FLD encoder to train low-level policy. 
The trained FLD encoder, through the propagation of latent dynamics, concurrently generates embeddings for global periodic parameters $\theta_t=(f_t, a_t, b_t)$, as well as for local phase states $\phi_t$ extracted from motion clips.
The global parameters $f_t$, $a_t$, and $b_t$ correspond to the frequency, amplitude, and offset of the latent trajectories. 
The parameterization leverages an autoencoder-like architecture to explicitly model the latent dynamics using a time-invariant frequency $f_t$ and a fixed time step $\Delta t$ with an autoencoder-like structure, as defined by:
\begin{equation}
    \mathbf{z}_t=(\theta_t, \phi_t) = \mathbf{enc}(\mathbf{s}_t), \quad \mathbf{\hat{z}}_{t+i} = (\theta_t,\phi_t+if_t\Delta t),
    \label{eq:fld_enc}
\end{equation}
\vspace{-0.5cm}
\begin{equation}
    \mathbf{\hat{s}}_{t+i} = \mathbf{dec}(\mathbf{\hat{z}}_{t+i}), \quad L^{FLD} = \sum_{i}^N \text{MSE}(\mathbf{s}_{t+i}, \mathbf{\hat{s}}_{t+i}).
    \label{eq:fld_dec}
\end{equation}
Here, $\mathbf{s}$ denotes the original motion sequence and $\mathbf{z}$ is latent representation, while $\mathbf{enc}(\cdot)$, and $\mathbf{dec}(\cdot)$ correspond to the encoding, and decoding operations, respectively. 
Our encoded state space comprises the base’s linear and angular velocities, the projected gravity vector in the robot base frame, and the joint positions.
The overall loss $L^{FLD}$ is computed as the reconstruction error between the reference and predicted state sequences over $N$ consecutive segments, effectively capturing the motion’s global periodic features.
For further details on the training hyperparameters and network architecture of the FLD model, refer to \tabref{tab:params_fld} to \tabref{tab:network_fld} and consult original FLD paper.

\subsection{Reward Setup for Low-Level and High-Level Policy Training} \label{sec:reward_low_high_detail}

\begin{figure}[h!]
    \centering
    \includegraphics[clip, trim=0cm 0cm 0cm 0cm,width=0.98\textwidth]{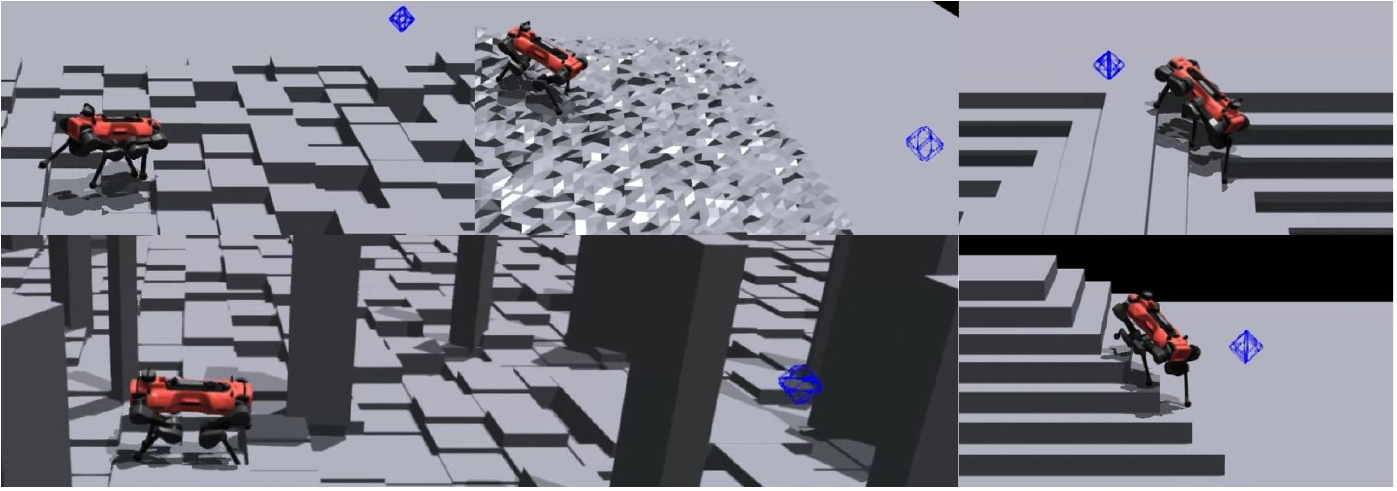}
    \caption{Overview of the training environment and terrain configuration for high-level policy. The setup includes diverse terrain types, boxes, stairs, rugged slopes, and high obstacles with the blue marker indicating the goal position.}
    \label{fig:overview_Terrains}
    \vspace{-0.4cm}
\end{figure}

The reward equations used for low-level policy and high-level policy are summarized in \tabref{tab:reward_lowlevel} to \tabref{tab:symbol}.

\begin{table}[h!]
    \centering
    \caption{Reward Equations for Low-Level Policy}
    \label{tab:reward_lowlevel}
    \begin{tabular}{l|l}
    \hline
    \textbf{Name} & \textbf{Equation} \\ \hline
    Linear Velocity Tracking & $2\exp{(-\norm{\mathbf{v}_{t}^b-\mathbf{v}_{t}^{b,ref}}^2)}$  \\
    Angular Velocity Tracking & $0.8\exp{(-0.8\norm{\mathbf{w}_{t}^b-\mathbf{w}_{t}^{b,ref}}^2)}$ \\
    Joint Position Tracking & $1.4\exp{(-2\sum_{i=1}^{12}(q_{t,i}-q_{t,i}^{ref})^2)}$ \\
    Projected Gravity Tracking & $0.8\exp{(-3\norm{\mathbf{g}_t-\mathbf{g}_t^{ref}}^2)}$ \\
    Action Rate & $-0.005\sum_{i=1}^{12}(a_{t,i}-a_{t-1,i})^2$ \\
    Collision & $-\sum_{k\in \text{thigh, shanks}} c_k$ \\ 
    Torque Limits & $-0.2\sum_{i=1}^{12}\text{max} (\tau_{t,i} - \tau_{lim}, 0)$ \\ 
    Torques & $-0.00002\sum_{i=1}^{12} \tau_{t,i}^2$ \\ 
    Joint Acceleration & $-0.00007\sum_{i=1}^{12}{\ddot{q}_{t,i}}^2$ \\ 
    Feet Acceleration & $-0.0001\sum_{i=1}^4 \norm{\mathbf{v}_{t,i}^{f} - \mathbf{v}_{t-1,i}^{f}}^2$ \\ 
    Contact Forces & $-0.005\sum_{i=1}^4{F_{t,i}^f}^2$ \\ 
    \hline
    \end{tabular}
    \vspace{-0.2cm}
\end{table}

\begin{table}[h!]
    \centering
    \caption{Reward Equations for High-Level Policy}
    \label{tab:reward_highlevel}
    \begin{tabular}{l|l}
    \hline
    \textbf{Name} & \textbf{Equation} \\ \hline
    Position Tracking & $15r_{reach}$ \\
    Heading Velocity & $5r_{vel}$ \\
    Joint Residual & $-0.1\sum_{i=1}^{12}(a^{res}_{t,i})^2$ \\
    Action Rate & $-0.005\sum_{i=1}^{12}(a^{res}_{t,i}-a^{res}_{t-1,i}+a_{t,i}-a_{t-1,i})^2$ \\
    Collision & $-\sum_{k\in \text{thigh, shanks}} c_k$ \\ 
    Stand Still & $-2.5\norm{\mathbf{v}^b_t}^2-\norm{\mathbf{w}_t^b}^2$ \\ 
    Stand Pose & $-0.2\sum_{i=1}^{12}(q_{t,i}-q_{i}^*)^2-5(g_{x,t}^2+g_{y,t}^2)$ \\ 
    Torque Limits & $-0.2\sum_{i=1}^{12}\text{max} (\tau_{t,i} - \tau_{lim}, 0)$ \\ 
    Termination & $-200$ \\ 
    Torques & $-0.00002\sum_{i=1}^{12} \tau_{t,i}^2$ \\ 
    Joint Acceleration & $-0.00007\sum_{i=1}^{12}{\ddot{q}_{t,i}}^2$ \\ 
    Feet Acceleration & $-0.0001\sum_{i=1}^4 \norm{\mathbf{v}_{t,i}^{f} - \mathbf{v}_{t-1,i}^{f}}^2$ \\ 
    Contact Forces & $-0.005\sum_{i=1}^4{F_{t,i}^f}^2$ \\ 
    \hline
    \end{tabular}
    \vspace{-0.2cm}
\end{table}

\begin{table}[h!]
    \centering
    \caption{Symbols for \tabref{tab:reward_lowlevel} and \tabref{tab:reward_highlevel}}
    \label{tab:symbol}
    \begin{tabular}{l|l}
    \hline
    \textbf{Symbol} & \textbf{Description} \\ \hline
    $\omega_{z}^b$ & Yaw base velocity \\
    $c_k$ & 1 if body $k$ is in contact, 0 otherwise \\
    $\mathbf{v}^b,\mathbf{v}^f_i$ & linear velocity vector of base and foot $i$ \\
    $\mathbf{w}^b$ & Angular velocity vector of base \\
    $q_{i} ,q_{i}^*,q_{i}^{ref}$ & actual, default and reference position of joint $i$ \\
    $\ddot{q}_{i}$ & acceleration of joint $i$ \\
    $\tau_{i}, \tau_{lim}$ & torque and torque limit of joint $i$ \\
    $F_i^f$ & Contact force of foot $i$ \\ 
    $\mathbf{g}, g_x, g_y$ & Projected gravity vector, projected gravity \\
     & along x and y axis in robot frame \\ \hline
    \end{tabular}
\end{table}

\subsection{Training Setup and Hyperparameters} \label{sec:hyper_low_high_detail}
Hyperparameters for FLD model, low-level policy, high-level policy training are presented in~\tabref{tab:params_fld} to \tabref{tab:params_student}. 
All simulations are performed in Isaac Gym \cite{makoviychuk2021isaac}. 
Both the low-level policy and the high-level teacher policy are trained in parallel environments using the PPO algorithm~\citep{schulman2017proximalpolicyoptimizationalgorithms} on a single NVIDIA RTX 4090. 
The low-level policy is trained exclusively on flat terrain, whereas the high-level policy is trained in the more complex environment illustrated in~\figref{fig:overview_Terrains}.
\begin{table}[h!]
    \centering
    \caption{Training Hyperparameters for FLD}
    \label{tab:params_fld}
    \begin{tabular}{l|l}
    \hline
    \textbf{Configuration} & \textbf{Values} \\ \hline
        Step Time Seconds & 0.02 \\
        Latent Channel & 4 \\
        Propagation Horizon & 30 \\ 
        Trajectory Segment Length & 31 \\ 
        Propagation Decay & 1.0 \\
        Learning Rate & 0.0001 \\ 
        Weight Decay & 0.0005 \\
        Number of Mini-Batches & 20 \\\hline
    \end{tabular}
\end{table}
\begin{table}[h!]
    \centering
    \caption{Network Architecture for FLD}
    \label{tab:network_fld}
    \begin{tabular}{l|l|l|l}
    \hline
    \textbf{Network} & \textbf{Layer} & \textbf{Output Size} & \textbf{Activation}\\ \hline
        Encoder & Conv1d & 64 $\times$ 31 & ELU \\ 
         & Conv1d & 64 $\times$ 31 & ELU \\ 
         & Conv1d & 4 $\times$ 31 & ELU \\ \hline
         Phase Encoder & Linear & 4 $\times$ 2 & Atan2 \\ \hline
         Decoder & Conv1d & 64 $\times$ 31 & ELU \\ 
          & Conv1d & 64 $\times$ 31 & ELU \\ 
         & Conv1d & 27 $\times$ 31 & ELU \\ \hline
         
    \end{tabular}
\end{table}

\begin{table}[h!]
    \centering
    \caption{Training Hyperparameters for PPO}
    \label{tab:params_ppo}
    \begin{tabular}{l|l}
    \hline
    \textbf{Configuration} & \textbf{Values} \\ \hline
    Actor and Critic Network & MLP \\
     & Hidden Layer Size (512, 256, 128) \\
    Robot Number & 4096 \\
    Step Number per Policy Update & 48 \\
    Entropy Coefficient & 0.002 \\
    Learning Rate & adaptive \\ 
    GAE-lambda & 0.95 \\
    Discount Factor & 0.99 \\
    Coefficient of KL Divergence & 0.01 \\
    Clip Ratio & 0.2 \\
    Mini Batch Size & 49152 \\ \hline
    \end{tabular}
\end{table}

\begin{table}[h!]
    \centering
    \caption{Training Hyperparameters for Student Policy}
    \label{tab:params_student}
    \begin{tabular}{l|l}
    \hline
    \textbf{Configuration} & \textbf{Values} \\ \hline
    Truncate Step for TBPTT & 15 \\
    Learning Rate & 0.0005 \\ 
    Number of Learning Epochs & 2 \\ 
    \hline
    \end{tabular}
\end{table}

\section{Additional Details for Experimental Results}
\subsection{Additional Results for Low-Level Policy}
\begin{figure}[h!]
    \centering
    \includegraphics[clip, trim=0cm 0cm 0cm 0cm,width=0.98\textwidth]{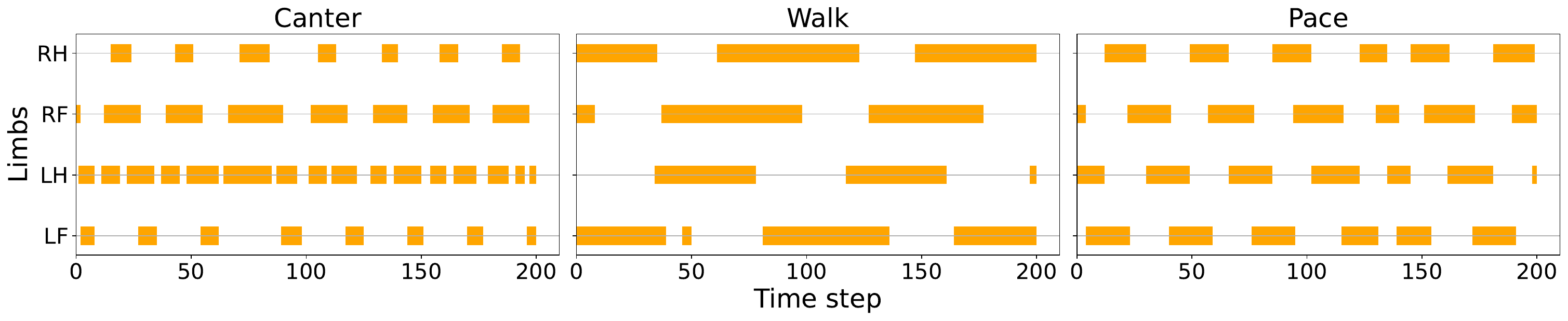}
    \caption{Footfall sequence of motion skills learned in low-level policy.}
    \label{fig:foot_sequence_low_level}
    \vspace{-0.3cm}
\end{figure}

\Figref{fig:foot_sequence_low_level} shows the footfall patterns of three flat-terrain skills in low-level policy under different latent embedding commands.

\subsection{Evaluation on High-Level Student Policy in Simulation} \label{sec:eval_student_policy_sim}

We evaluate the high-level student policy with noisy observation in simulation and report its success rate for each terrain in Table \ref{tab:succ_rate_student}. For each terrain type, we run 100 trials with the goal placed 5 meters from the robot’s start in a random direction.

\begin{table}[h!]
    \centering
    \caption{Success Rate of High-Level Student Policy in Simulation}
    \label{tab:succ_rate_student}
    \begin{tabular}{l|l}
    \hline
    \textbf{Terrain Type} & \textbf{Success Rate} \\ \hline
    0.25m Stairs (Up) & 84/100 \\
    0.25m Stairs (Down) & 85/100 \\
    24-deg Slope (Up) & 90/100 \\ 
    24-deg Slope (Down) & 95/100 \\ 
    Random Boxes & 81/100 \\ 
     Random Boxes with High Obstacles & 75/100 \\ 
     Flat ground with High Obstacles & 90/100 \\ 
     \hline
    \end{tabular}
\end{table}

\subsection{Simulation Setup for Evaluation on Joint Residuals}
\begin{figure}[h!]
    \centering
    \includegraphics[clip, trim=0cm 0cm 0cm 0cm,width=0.98\textwidth]{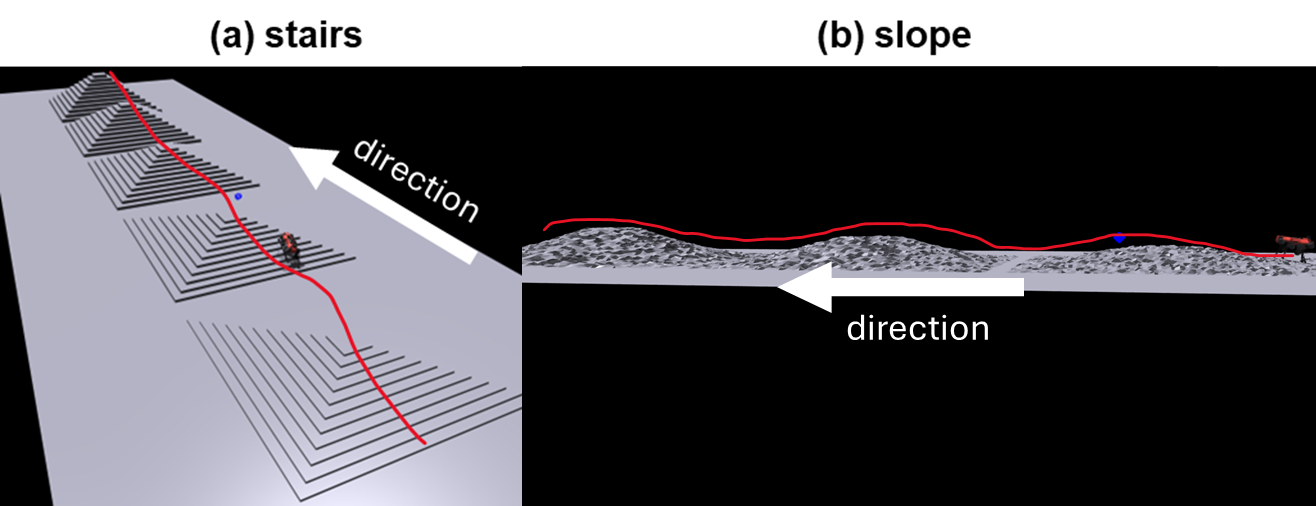}
    \caption{Overview of (a) pyramid stairs and (b) pyramid slope terrains with progressively increasing difficulty levels during inference. The blue marker indicates the goal position at the current frame which is moving from easier to more challenging terrain level, while the white arrow denotes the robot's walking direction. The red line traces an example trajectory of the robot’s base position.}
    \label{fig:residual_terrain_levels}
    \vspace{-0.3cm}
\end{figure}

As described in \secref{sec:highlevel}, we investigate how joint residuals vary across two terrain types, pyramid stairs and pyramid slopes, as the terrain difficulty progressively increases.
The corresponding simulation setup is illustrated in \figref{fig:residual_terrain_levels}.

\subsection{Additional Comparative Analysis on Energetic Efficiency}
\begin{figure}[h!]
    \centering
        \includegraphics[clip, trim=0cm 0cm 0cm 0cm,width=0.55\textwidth]{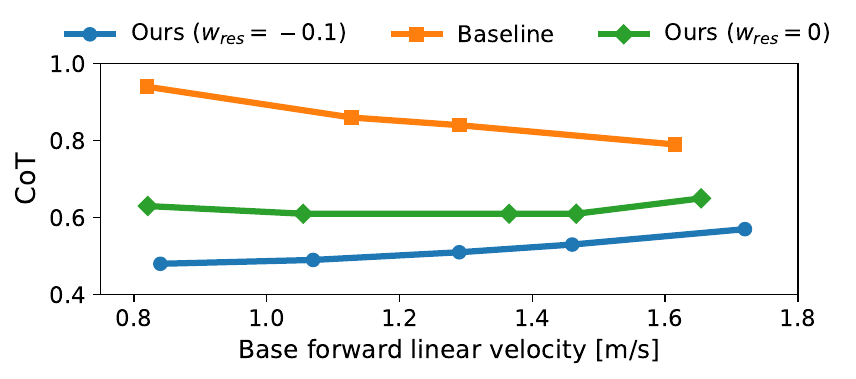}
    \caption{Energy efficiency comparison among three models: (i) the baseline RL model trained from scratch, (ii) our proposed framework trained with motion priors and a proper joint residual penalty ($w_{res}=-0.1$) (iii) a variant of our model without residual penalization ($w_{res}=0$). The evaluation spans a wide range of forward velocities achieved via varying maximum velocity commands $v_{cmd}$. The CoT is averaged over a 20-second period of continuous forward motion.}
    \label{fig:baselineVSOurs}
    \vspace{-0.5cm}
\end{figure}

Building on the results in \secref{sec:com_baseline}, we perform a comparative analysis of energetic efficiency of our models and the baseline under similar reward settings.
In this experiment, each model continually walks forward by following a moving goal position.
\Figref{fig:baselineVSOurs} illustrates the Cost of Transport (CoT) on flat ground for the baseline and our two models, measured as the robot walks at varying speeds. 
CoT is calculated using the following equation.
\begin{equation}
    CoT = \frac{\sum_i^{12}|\tau_{t,i}\dot{q}_{t,i}|}{mgv_{b}},
\end{equation}
where $m$ denotes the total mass of the robot, $g(=9.8 \mathrm{m/s^2})$ the gravity constant, $v_b$ the forward linear base velocity, $\dot q_{t,i}$ the velocity of joint $i$.

Although our framework does not explicitly optimize the energy consumption in the reward terms, our two models trained with motion priors achieve a lower CoT. This likely reflects the baseline model’s insufficient constraints on vertical motion, whereas the incorporation of motion priors could naturally avoid those jumping gaits. Additionally, our comparison shows that removing residual regularization leads to a slight increase in CoT, probably due to the unconstrained residuals.

These findings imply that the animal-like motion priors can have the potential to steer training toward energy efficient gaits. 
However, further research is still needed to quantify this benefit and compare it against explicitly adding energy penalties to the reward functions.

\subsection{Training High-Level Policies with Minimal Reward Terms}
\begin{table}[t!]
    \centering
    \caption{Reward Equations for High-Level Policy in Extended Experiments}
    \label{tab:reward_highlevel_ext}
    \begin{tabular}{l|l}
    \hline
    \textbf{Name} & \textbf{Equation} \\ \hline
    Position Tracking & $15r_{reach}$ \\
    Heading Velocity & $5r_{vel}$ \\
    Joint Residual & $-0.5\sum_{i=1}^{12}(a^{res}_{t,i})^2$ \\
    Collision & $-\sum_{k\in \text{thigh, shanks}} c_k$ \\ 
    Termination & $-200$ \\ 
    \hline
    \end{tabular}
    \vspace{-0.2cm}
\end{table}
To further investigate how effectively our framework reduces the effort required for reward shaping, we trained an additional high-level policy using a minimal set of reward terms (\tabref{tab:reward_highlevel_ext}), employing the same low-level motion priors. 
As illustrated in the supplementary videos, the new policy also successfully achieves perceptive locomotion and local navigation using animal-like gaits across varied terrains, despite the minimal reward configuration. 
Notably, unlike the model shown in the main paper, this simplified reward setup leads the policy to adopt a cantering gait rather than a walking gait. 
With the integration of learned joint residuals, the new high-level policy can still effectively traverse challenging terrains while maintaining the cantering gait. 
These results further confirm that our framework can efficiently produce high-level policies with natural, animal-like locomotion and local navigation across diverse terrains, all while reducing reward complexity and tuning effort.

\end{document}